\documentclass{article} % For LaTeX2e
\usepackage{nips15submit_e,times}
\usepackage{hyperref}
\usepackage{url}
\usepackage{pdfpages}
\usepackage{graphics}
\usepackage{xcolor}
\usepackage{enumitem}
\usepackage{amsmath}
\usepackage{subfig}

\title{AAD: Adaptive Anomaly Detection through traffic surveillance videos}

\author{
Mohammad Farhadi Bajestani \\
\texttt{mfarhadi@asu.edu} \\
\And
Seyed Soroush Heidari Rahmat Abadi \\
\texttt{Soroush.Heidari@asu.edu} \\
\And
Seyed Mostafa Derakhshandeh Fard \\
\texttt{smd.fard@asu.edu} \\
\And
Roozbeh Khodadadeh \\
\texttt{rkhodada@asu.edu} 
}

\nipsfinalcopy % Uncomment for camera-ready version

\begin{document}

\maketitle

\begin{abstract}
Anomaly detection through video analysis is of great importance to detect any anomalous vehicle/human behavior at a traffic intersection. While most existing works use neural networks and
conventional machine learning methods based on provided dataset, we will use object recognition
(Faster R-CNN) to identify objects labels and their corresponding location in the video scene as
the first step to implement anomaly detection. Then, the optical flow will be utilized to identify
adaptive traffic flows in each region of the frame.
Basically, we propose an alternative method for unusual activity detection using an adaptive anomaly detection framework. Compared to the baseline method described in the reference paper, our method is more efficient and yields the comparable accuracy. 
\end{abstract}

\section{Introduction}
Anomaly detection is usually defined as the set of techniques and methods in machine learning and data mining that can detect an unusual or non-conforming pattern in the data. Anomaly detection is of special importance in image recognition because of the ubiquity of surveillance systems and the emerging of self-driving cars which heavily rely upon anomaly detection for correct operation. Obviously, the objective is to reduce the role of humans in detection of anomalies to a minimum and at the same time have the capability to detect anomalies no matter how rare they are.

One of the challenges in defining anomalies in the context of video surveillance is that in a non-stationary process, the concept of normality itself changes. Entry of a new object into the view is an anomalous event in the first few moments. After the first few moments, it becomes a normal event from the system's viewpoint. To address this problem, We will present an adaptive method to deal with the non-stationary nature of the problem. 

Another challenge is that anomalous events are rare. We will also propose a method to deal with the sparsity of the data.

In short, our method is a combination of supervised and unsupervised learning, is based on analyzing a spatial signal, uses blocks to eliminate noise (Which would otherwise happen if we use only pixels), is capable of lossless representation the video sequence, and uses variance as the criteria to measure deviation from normality.

\subsection{Anomaly Detection}

Anomaly detection is a broad topic with many applications in various fields. Here, we confine our discussion of anomaly detection to its application in video surveillance.

The definition of anomaly is associated with our understanding of the normal state. For example, players running in a football field is considered normal activity, while people running in a parade is an instance of unusual activity \cite{Popoola2012}. One immediate consequence of this notion is that it is very easy for human observers to identify anomalies because they usually have a very good understanding of the context of the video. We need to come up with a way to define an anomaly for computers. For our purposes we define anomalies as temporal or spatial outliers which means "an action done in an unusual time or in unusual location"\cite{Popoola2012}.

With this view of anomaly, it seems that clustering is one natural choice for detecting anomalies. We can cluster features together. Anomaly is detected if there are objects that are not contained in any of our "normal" clusters. This is the basis of the technique used in the paper written by Li et al\cite{Lee_2015}.

To successfully detect anomalies, any proposed method needs to distinguish between empty and crowded scenes and between local and global anomalies. Behavior classification might also be needed to enable the algorithm to differentiate the normal from anomalous state.

\subsection{Object Recognition}

Object recognition is defined as the technology for finding and identifying objects in an image or video sequence\cite{wiki:ObjectRec}. Another way of thinking about neural networks is that they are function approximation machines\cite{Goodfellow-et-al-2016}. Going back to the discussion on kernel tricks, neural networks differ from traditional models in that they \textit{learn} the kernel transform $\Phi$ compared to using a generic or a manually engineered kernel function. Image recognition mostly employs Convolutional Neural Networks (CNN). These networks use the convolution operation in place of matrix multiplication in at least one of their layers.
To recognize objects, we first need to detect images. The first stage is usually an edge detection layer. Edge detection can be done by subtracting pixel values from neighboring pixels. The result is then passed through a pooling layer. Pooling layer replaces a pixel's value (or a block) with the summary statistics of its neighbors. The used statistics can be maximum, average or any other valid summary statistic function. Convolution in general is an expensive operation and running it on whole images can be very computationally expensive.

The state of the art in the image recognition field is to use a subset of neural networks called Regions with Convolutional Neural Networks (R-CMN). Following the previous paragraph, the next step in R-CNN is image segmentation which divides image into separate regions that may be of interest. The algorithm will also give bounding boxes around each region.

R-CNN is still slow. Researchers have developed Fast-RCNN and Faster-RCNN which perform considerably faster compared to RCNN. We use Faster-RCNN in our project to recognize and detect images. Despite its limits, faster R-CNN adds a significant improvement to the efficiency of the our general algorithm.

\section{Related Works}
Our reference paper\cite{Lee_2015}, uses a motion influence map(MIM) based on optical flow. Optical flow is the technique to recover three dimensional motion from a sequence of time-ordered two-dimensional images\cite{beauchemin1995computation}. Optical flow allows the estimation of projected two dimensional image motion as either instantaneous image velocities or discrete image displacements\cite{beauchemin1995computation}. It is a convenient motion representation method.

The reference paper then builds the MIM. MIM is based on the idea that a moving object can influence nearby blocks in two ways: 1)motion direction, and 2)motion speed. We can calculate all such influences and create a map in which each block represents the quantized motion orientation of the block.

Our contribution to the reference paper is twofold. First, we use Fast-RNN image recognition to detect and categorize objects that will later help us to detect the type of anomaly. Second, instead of a motion influence map and clustering, we use a Gaussian distribution function that represents the level of anomaly in each block. We use the output of our faster-RCNN object recognition algorithm to assign object probabilities to various regions in the image.

\section{Adaptive Anomaly Detection (AAD)}
\begin{figure}[ht]
	\begin{center}
	\includegraphics[width=12cm]{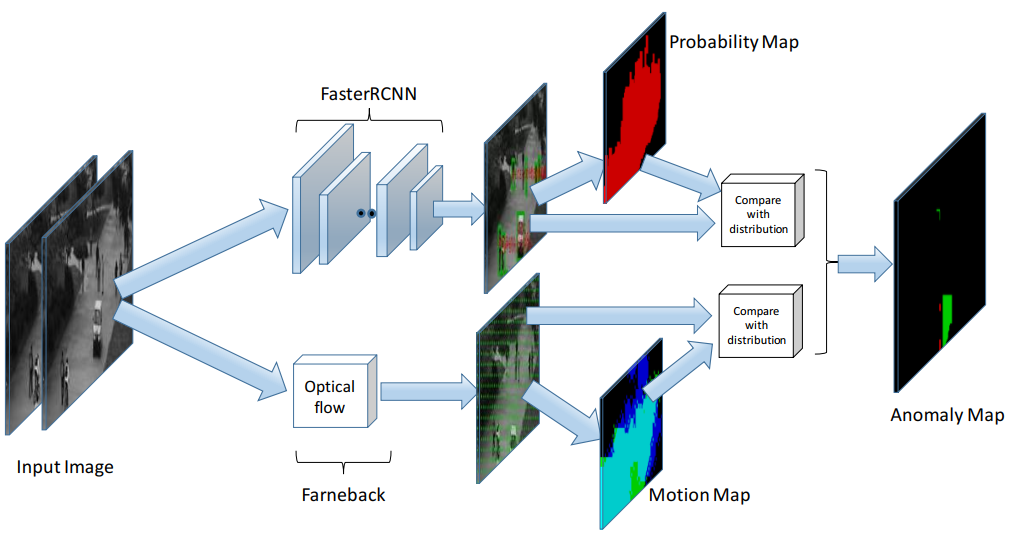}
    \caption{Complete Adaptive Anomaly Detection Flow}
    \label{fig:aad}
	\end{center}
\end{figure}

This section first introduces how to detect and localize anomalies via motion characteristics estimation over consecutive frames. After that, the object recognition algorithm applied on original data and the distribution map construction will be elaborated. Finally, the proposed adaptive anomaly detection approach will be explained in details. Figure \ref{fig:aad} shows the complete Anomaly Detection algorithm.

\subsection{Optical Flow}
In this paper, we observe different motion patterns at the block-level to spot unusual activities within a crowded scene. The motion patterns is estimated by optical flow using Farneback method. %\ref{} 
Optical flow is the pattern of apparent motion caused by brightness changes. Given two subsequent frames, the optical flow method tries to calculate how quickly the pixel is moving between the two frames which are taken at times $t+ \Delta t$ and what is the movement's direction. One of main limitations of optical flow is that it fails to extract motion characteristics from static or very slow moving objects. Also, based on our  primary results, pixels' displacements between consequent frames wasn't significant. Therefore, we picks every other frames to compare with and generate the optical flow's output. Initial output is a 3D matrix that has displacement information of each pixel across the image in both $(x,y)$ directions, that basically points to where that pixel can be found in another frame.  

% \begin{picture} (100,150)
% \put(25,10){\includegraphics[width=6cm]{op1.png}}
% \put(40,105){$I(x,y)$}

% \put(225,10){\includegraphics[width=6cm]{op2.png}}
% \put(245,120){$I(x + \Delta x,y + \Delta y)$}
% \end{picture} \\

\begin{figure*}[ht]
\centering
  \subfloat[]
  {\includegraphics[width=0.45\textwidth]{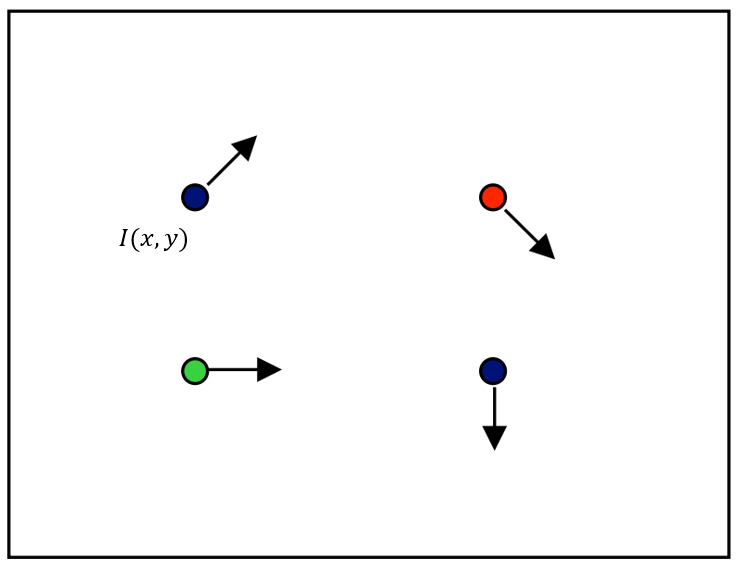}\label{op1}}
  \hfill
  \subfloat[]
  {\includegraphics[width=0.45\textwidth]{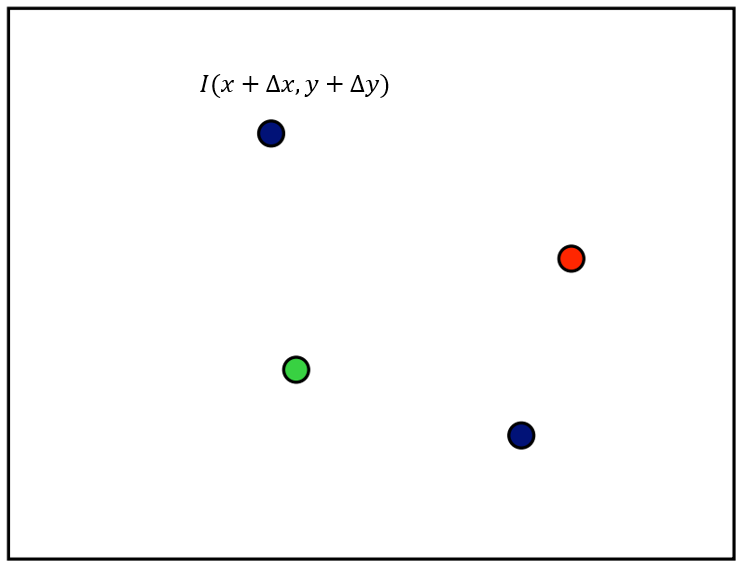}\label{op2}}
  \caption{Pixel movement across the frame after $\Delta t$}
\end{figure*}
\begin{figure}
	\begin{center}
	  	\includegraphics[width=5cm]{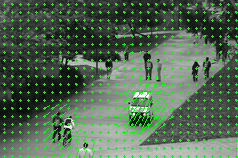}    
  		\includegraphics[width=5cm]{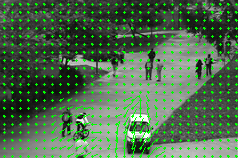}
  		\caption{Optical Flow Outputs.}
  		\label{fig:opticalflow}
	\end{center}
\end{figure}

\clearpage
Given two frames at times $t-2$ and $t$, a voxel at location $(x,y)$ with intensity $I(x, y)$ will have moved by $\Delta x, \Delta y$ between the two image frames with $\Delta t$ time difference. The Brightness Constancy equation is as follows: 
\begin{equation}
I(x,y,t) = I(x+ \Delta x, y + \Delta y, t + \Delta t) 
\end{equation} 
Then it can written as:
\begin{equation}
I(x+ \Delta x, y + \Delta y, t + \Delta t) = I(x,y,t) + \frac{\partial I}{\partial x}\Delta x +
\frac{\partial I}{\partial y}\Delta y +
\frac{\partial I}{\partial t}\Delta t 
\end{equation}
From these equations it follows that:
\begin{equation}
\frac{\partial I}{\partial x}\frac{\Delta x}{\Delta t} +
\frac{\partial I}{\partial y}\frac{\Delta y}{\Delta t} +
\frac{\partial I}{\partial t}\frac{\Delta t}{\Delta t} = 0 
\end{equation}
which can be transformed to:
\begin{equation}
\frac{\partial I}{\partial x}V_x +
\frac{\partial I}{\partial y}V_y +
\frac{\partial I}{\partial t} = 0 
\end{equation}
Thus, the Optical Flow equation is stated as:
\begin{equation} \label{eq:5}
I_xV_x + I_yV_y = -I_t 
\end{equation} 
The equation \ref{eq:5} cannot be solved with two unknown variables $I_x$ and $I_y$. Therefore, several optical flow methods introduce additional conditions to calculate the actual flow and one of them is Gunner Farneback's algorithm which is explained in \cite{Farneback}. We have used OpenCV library that is equipped with Gunner Farneback's algorithm.\\
After estimating the pixel-wise dense optical flow, we process the data through two pooling stages in order to remove the noise and reduce the processing time. First, we apply $2*2$ average pooling with two major objective which are noise and dimensionality reduction that consequently results in improved speed and performance. Then, we kept decreasing the dimension by set the maximum change to be the the block representative via a $2*2$ max pooling stage. Figure ~\ref{fig:opticalflow} shows that Optical flow clearly is able to detect the car as an anomaly due to its speed difference with pedestrians.

\subsection{Object Recognition}
This object recognition is our complementary tool to expose and localize the anomalies within a packed and crowded scene. We have utilized Faster R-CNN that is an object recognition algorithm to find unusual objects in our surveillance videos. Faster R-CNN is equipped with Region Proposal Network (RPN) in order to rapidly and efficiently process the whole frame and determine the region of interest that needs further processing. The input image goes through a network of convolution layers and transforms to a convolutional feature map. Next, The RPN uses the computed a feature map to detect a predefined number of regions (bounding boxes), that may contain objects.

\begin{figure}[ht]
	\begin{center}
	  	\includegraphics[width=12cm]{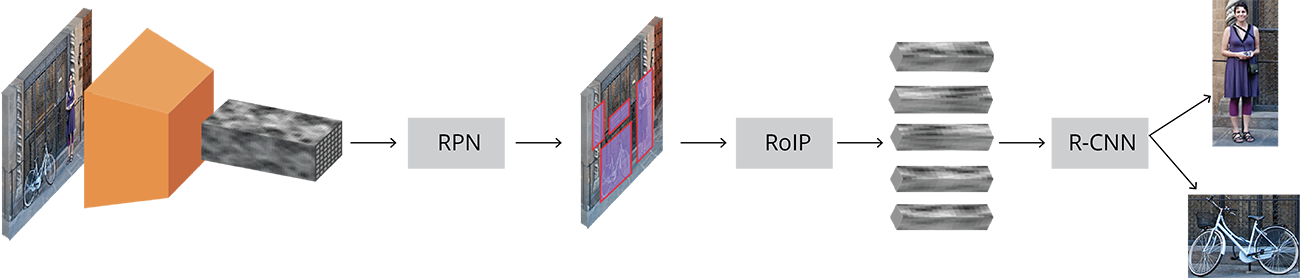}    
  		\caption{Faster R-CNN Architecture.}
  		\label{fig:FasterRCNN}
	\end{center}
\end{figure}

The bounding boxes is used as the input to the Region of Interest (ROI) pooling layer, which performs a pooling operation on a part of the input map that corresponds to region proposals in the original image. Finally, the R-CNN module either classifies the content of the bounding boxes or it adjust the bounding boxes coordinates to better fit the content and consequently can be classified precisely. 

\subsection{Distribution Map}
In the proposed Distribution Map, we introduce a 5-tuple that includes pixel-wise mean, max, min, variance, and number of frames of video sequences. We build two distribution maps based on both optical flow and object recognition outputs.Then, the distribution map is used to expose pixel's values that shows drastic changes $(\mu \pm k\sigma)$ compared to velocity distribution or is classified as an unusual object in comparison with object's history in that area of the frame. Also, We update the computed temporal 5-tuple of video sequences continuously after each change in pixel's value. In order to calculate the temporal mean and variance over streaming video frames, each time that the new data is loaded and differs from previous values, the new 5-tuple is computed. The $\mu$ is updated based on incoming data as follows:
\begin{equation}
\mu_{N} = \frac{\mu_{N-1}(N-1) + x_{N}}{N}
\end{equation}
and then if we detect an anomaly in the video sequences, we reduce by a half the number of frames, so the weight of the incoming data will be larger. That means, the algorithm will adapt itself and stop detecting false positive anomaly.Therefore the updated mean after anomaly detection is calculated in this way:
\begin{equation}
\mu_{N} = \frac{\mu_{N-1}\frac{(N-1)}{2} + x_{N}}{\frac{N-1}{2}+1}
\end{equation}
Next we calculated the $\sigma^2$ as follows.
\begin{equation} \label{eq:7}
\sigma_N^2 = \frac{(N-2)\sigma_{N-1}^2 + (x_N-\bar{x}_N)(x_N - \bar{x}_{N-1})}{N-1}
\end{equation}
The Equation \ref{eq:7} can be derived from Welford's method as described below. Welford’s method is a usable single-pass method to calculate the variance. It is quite surprising and computationally efficient that how simple the sums of squared differences for $N$ and $N-1$ samples results in the new variance\cite{Ling}.
\begin{align} 
&(N-1)\sigma_N^2 – (N-2)\sigma_{N-1}^2 \\ 
&= \sum_{i=1}^N (x_i-\bar{x}_N)^2-\sum_{i=1}^{N-1} (x_i-\bar{x}_{N-1})^2 \\ 
&= (x_N-\bar{x}_N)^2 + \sum_{i=1}^{N-1}\left((x_i-\bar{x}_N)^2-(x_i-\bar{x}_{N-1})^2\right) \\ 
&= (x_N-\bar{x}_N)^2 + \sum_{i=1}^{N-1}(x_i-\bar{x}_N + x_i-\bar{x}_{N-1})(\bar{x}_{N-1} – \bar{x}_{N}) \\ 
&= (x_N-\bar{x}_N)^2 + (\bar{x}_N - x_N)(\bar{x}_{N-1} - \bar{x}_{N}) \\ 
&= (x_N-\bar{x}_N)(x_N - \bar{x}_N - \bar{x}_{N-1} + \bar{x}_{N}) \\ 
&= (x_N-\bar{x}_N)(x_N - \bar{x}_{N-1}) \\ 
\end{align}

\section{Experimental Results and Analysis}
\label{others}

\subsection{Dataset Description}
To evaluate the performance of the proposed method, we put to the test on several dataset, i.e., the PASCAL VOC, UMN, and USDC, which contain global and local anomalies. 
\begin{enumerate}[label=\Alph*.]
\item \textit{PASCAL Visual Object Classes (VOC)}\footnote{Available at http://host.robots.ox.ac.uk/pascal/VOC/}\\ \newline
The VOC challenge is a benchmark in visual object category recognition and detection, providing the vision and machine learning communities with a standard dataset of images and annotation, and standard evaluation procedures. Organized annually from 2005 to 2012, the challenge and its associated dataset has become accepted as the benchmark for object detection. In VOC 2012, the train/validation data (1.9 GB) has 11,530 images containing 27,450 ROI annotated objects and 6,929 segmentations. The goal of PASCAL VOC is to recognize objects from a number of visual object classes in realistic scenes (i.e. not pre-segmented objects).  It is fundamentally a supervised learning problem in that a training set of labeled images is provided. It contains twenty object classes as shown in figure \ref{fig:pascal}. For each of the twenty classes, VOC predict the bounding boxes of each object of that class in a test image, with associated real-valued confidence. We use PASCAL VOC to train faster RCNN and detect objects, in our work. \cite{Everingham15} \\
\begin{figure}[ht]
	\begin{center}
    \includegraphics[width=9cm]{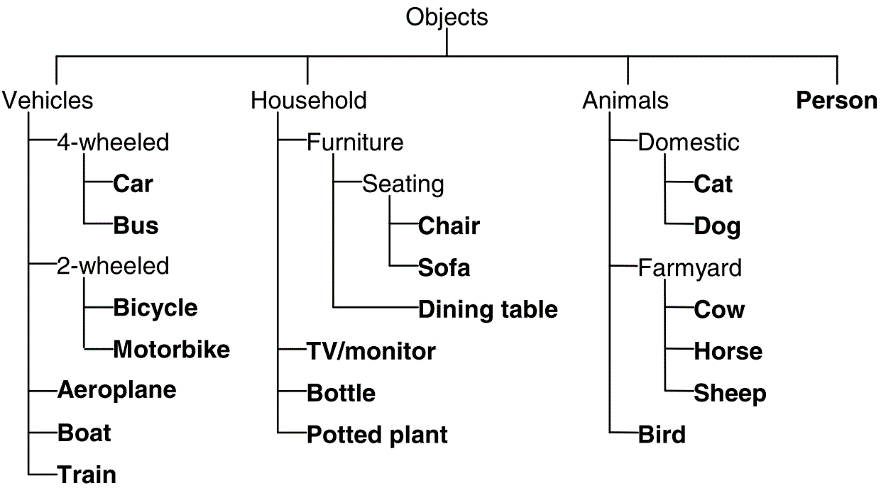}
    \caption{PASCAL Visual Object Classes}
    \label{fig:pascal}
	
	\end{center}
\end{figure}
\item \textit{UMN Dataset}\footnote{Available at http://mha.cs.umn.edu/Movies/Crowd-Activity-All.avi} \\ \newline
We validated the effectiveness of the proposed method on UMN dataset. It consists of 11 video clips of crowded escape scenarios from three different indoor and outdoor scenes. It includes 7740 frames in total (1450, 4415 and 2145 for scenes 1–3, respectively), where the frame size is 320 × 240. The normal events are pedestrians walking randomly on the square or in the mall, and the abnormal events are human spread running at the same time. \\
\item \textit{UCSD Dataset}\footnote{Available at http://www.svcl.ucsd.edu/projects/anomaly} \\ \newline
There are two sets of video clips in the UCSD dataset: Ped1 and Ped2. Figure (\ref{ped1}\ref{ped2}) shows sample frames Ped1 and Ped2 dataset. The dataset provides both frame- and pixel-level ground truths, which localize the regions where the unusual activities occur. Ped1 consists of 34 training clips and 36 test clips, where the frame size is 238×158. Ped2 consists of 16 training clips and 12 test clips, where the frame size is 360 × 240. The training clips include only normal activities of people walking along a pathway. For the test clips, unusual activities such as bicycling, skating, and driving a cart are taking place. 

\begin{figure*}[t]
\centering
  \subfloat[]
  {\includegraphics[width=0.45\textwidth]{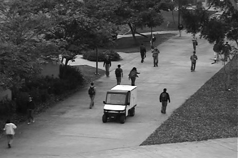}\label{ped1}}
  \hfill
  \subfloat[]
  {\includegraphics[width=0.45\textwidth]{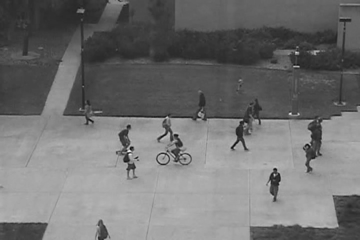}\label{ped2}}
  \caption{(a) UCSD Ped1 and (b) UCSD Ped1 sample frames}
\end{figure*}

\end{enumerate}

\subsection{Framework}
This project consists of two main feature extraction which needs different framework to implement. object recognition part has been implemented using PyTorch \cite{pytorch} and optical flow has been implemented using OpenCV\cite{opencv}.

PyTorch is an open source library in python for running machine learning algorithm \cite{pytorch}. This library has been implemented based on Torch. It is primarily developed by Facebook's artificial-intelligence research group. This library provide two high-level features, Tensor computation with GPU acceleration and deep neural networks. 

OpenCV (Open Source Computer Vision Library) is an open source library to run machine learning and computer vision algorithm. This library has interface for C++, java, Python and MATLAB. This library best option to run image processing algorithm such as optical flow which has been used in our project.

All modules of this project has been implemented in python.An Ubuntu machine with two Xeon CPUs, 64 GB RAM and one NVIDIA TITAN-X PASCAL has been used in this project.  
\begin{figure}[ht]
	\begin{center}
	  	\includegraphics[width=4.8cm]{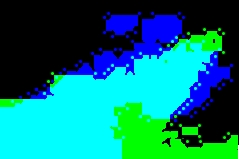}    
  		\includegraphics[width=5.2cm]{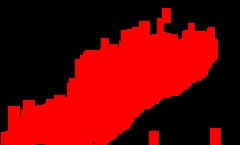}
  		\caption{Motion Distribution Map and Object probability Map (Just Person probability) in UCSD dataset}
  		\label{prob_map}
	\end{center}
\end{figure}
\subsection{Distribution Map}
In figure \ref{prob_map}, we show the motion distribution of UCSD in left figure. In this figure we show the mean of dx and dy which calculated frame by frame, dx represented by green color and dy represented by blue. In positions which one of the mean is significantly bigger than other one, we have dark blue or green. In middle of frame which we have motion to all direction we are more close to bright blue. black area represent area which we didn't observe any movement.

Next figure in \ref{prob_map} shows the probability map for one specific class. In this figure we select person as target class and show the existence probability of this class in each pixel with read color in UCSD dataset. It is clear that in region we have higher chance of observing human in that region we have very high probability to observe human. Dark region shows places which probability of observing human is close to zero. In real case, we have a tensor with size of image which each pixel has 22 variable that 21 of these keep number of observation for specific object and last total number of observed objects. In figure \ref{prob_map} we just shows the result for human class. 

Combination of these create our distribution for UCSD dataset. In our we create same distribution for UMN dataset. 

\subsection{Parameter Settings}
\textbf{Object Recognition module:} For object recognition module we are using FasterRCNN \cite{fasterrcnn}. We are using ResNet 101 as feature extractor for FasterRCNN which has been pre-trained over ImageNet. The initial learning rate has been set to 0.001, for learning decay we gamma learning decay with 0.1 decay ratio. The model has been trained over Pascal Dataset. The object selection threshold has been set to 0.8, this means objects detected with probability lower than 0.8 will be removed from our consideration map. The model has been trained in 20 epochs with 10000 iteration. model with lowest error rate over test has been selected for our adaptive anomaly detection system.

\textbf{Optical flow module:} For this module we are using Gunnar Farneback's algorithm in OpenCV. To get better result we compare two last frame with each other. In optical flow function we set number of pyramids to 3, averaging windows to 15, number of iteration to 5 and standard deviation of the Gaussian to 1.2.

\textbf{Hyper parameters:} we are creating a window over distribution and if our incoming data is inside we would classify it as normal. Otherwise it would be treated as an anomaly. The window size is equal to $k\sigma$ centered at $\mu$. The generated figure from both datasets stem from the fact that changing the window size $k$ results in significant True Positive and False Positive variation in final results. In this work, we modified our model's $k$ values with a range from 1 to 6. 

\begin{figure*}[ht]
\centering
  \subfloat{\includegraphics[width=0.5\textwidth]{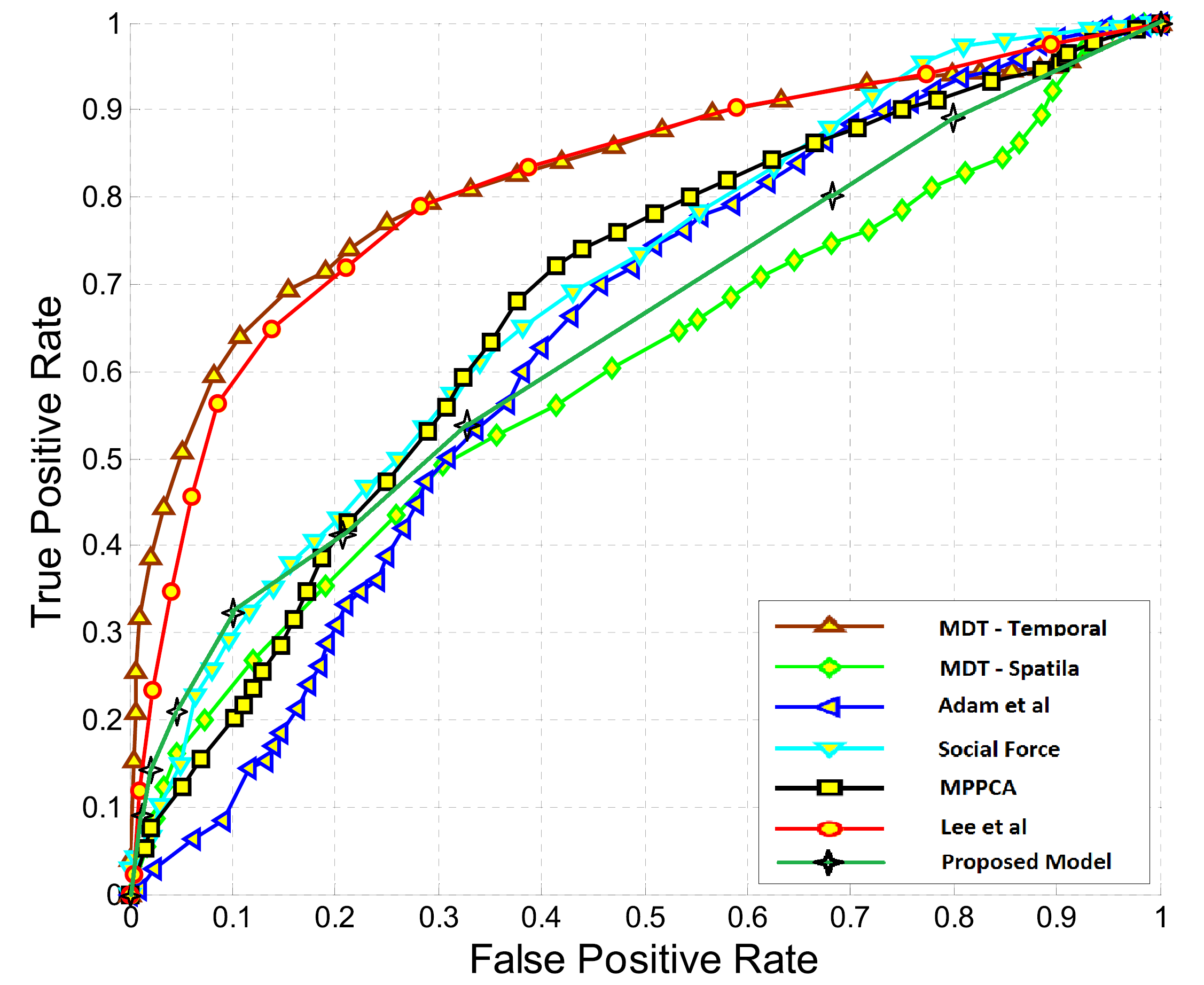}\label{ped1p}}
  \hfill
  \subfloat{\includegraphics[width=0.5\textwidth]{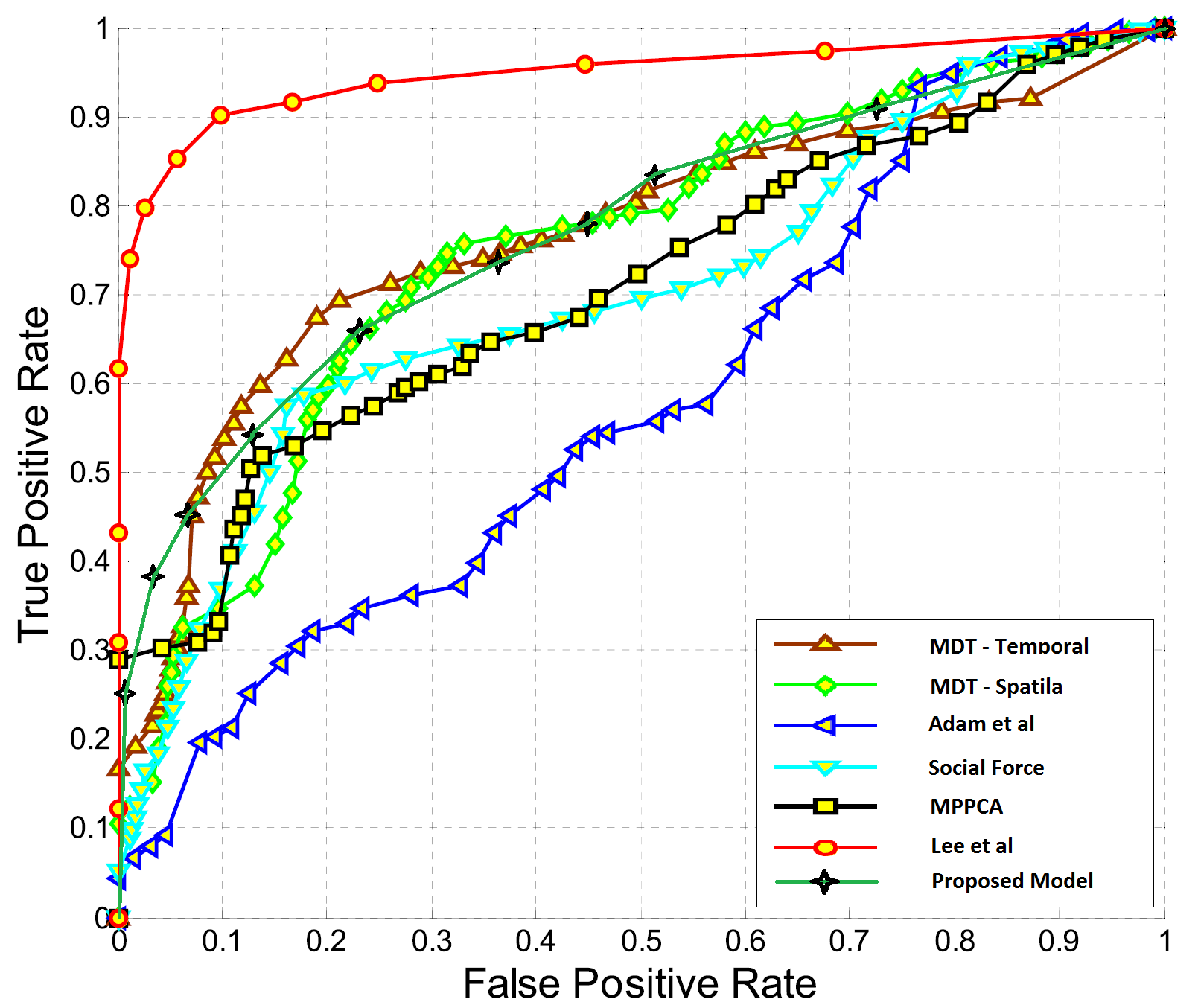}\label{ped2p}}
  \caption{(a) UCSD Dataset - Ped1 (b) UCSD Dataset - Ped1}
\end{figure*}
\subsection{Performance Evaluation}
Similar to the approach taken in the reference paper\cite{Lee_2015}, We also use Receiver Operating Characteristic (ROC) curves to evaluate performance. ROC curves show the trade-off between hit rates and false alarms\cite{FAWCETT2006861}. The idea is to map classification rates to a 2 by 2 matrix (for a binary classification algorithm) called \textit{confusion matrix}. This matrix shows the number of true as well as false positives and negatives. An ROC graph takes this matrix and gives a two dimensional box plot in which the horizontal axis is the false positive rate and the vertical axis is the true positive rate. In this graph the point (0,1) corresponds to perfect classification (no false positives) and point (1,1) corresponds to arbitrary assignment of class (a lot of false positives). We plotted our model performance based on different window size $k$. Note that the objective is getting large true positive values and small false positive values, that means we are interested in getting results close to upper left corner of the plot. Figure (\ref{ped1p},\ref{ped2p}) shows that if we set the $k=1$, we will end up with both false positive and true positive equal to 1. Also, if we set the value to the other extreme which is $k=6$ because the model does not detect any object as anomaly, both values of false positive and true positive tends to zero. Also, the performance difference between two dataset can caused by dissimilar point of view and variation in object's velocity (both speed and its direction). Keep in mind that the plotted performance is obtained without considering object recognition module's output. Thus, if we add the object recognition module to the test model, we experience slightly better performance (True Positive $=0.56$, Flase Positive $=0.36$ for Ped1). The reason behind poor performance of the object recognition module is that the Faster R-CNN is not trained over this dataset.  
\begin{figure}[ht]
\centering
	\includegraphics[width=12cm]{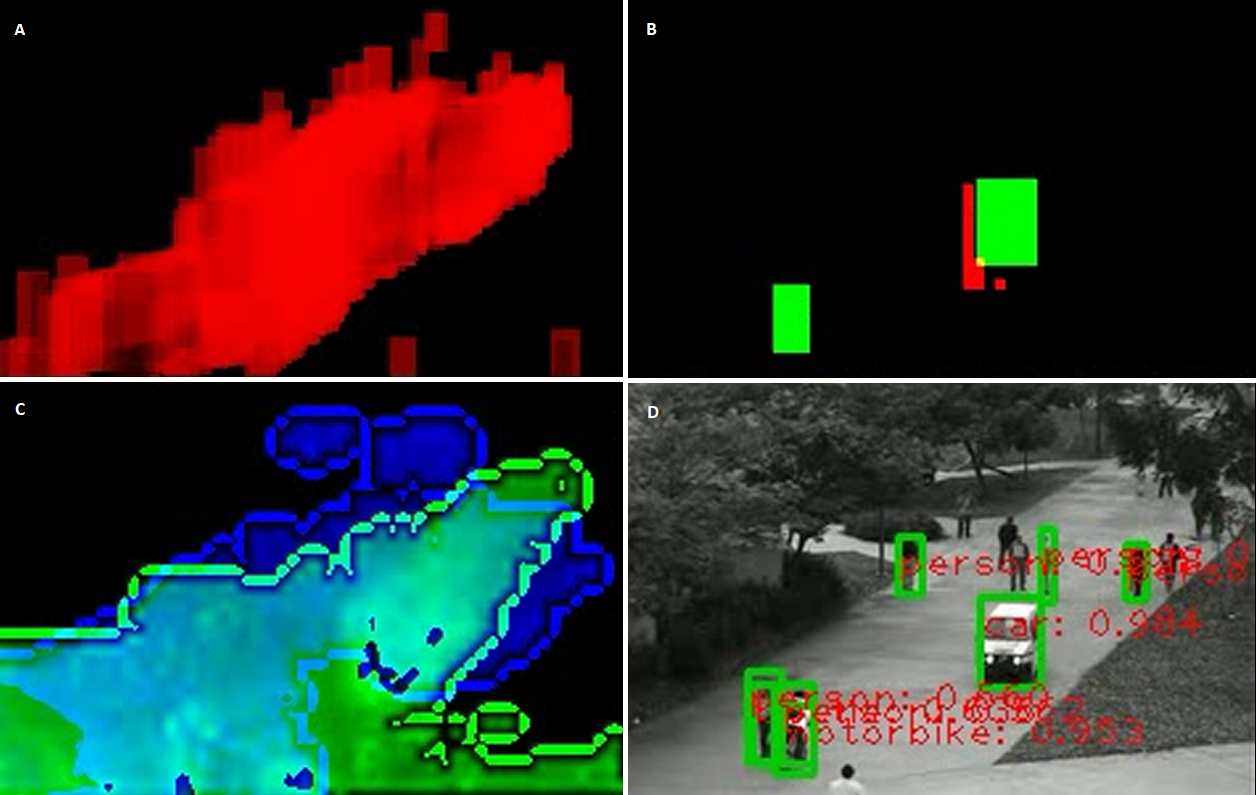}
  \caption{(a) Human Probability (b) Anomaly Map (c) Motion Distribution Map (d) Input Frame + Object Recognition Output}
  \label{screenshot}
\end{figure}

Also, Figure \ref{screenshot}b shows that the presence of the car in the street is correctly detected based on optical flow distribution map, also the dark shadow over the red-colored region in fig. \ref{screenshot}a support the same result. Although, we didn't meet the state of art accuracy, we achieved decent results compared to previous works. More importantly, we gained a extremely faster ($~2$) running time due to simpler model complexity

\section{Conclusion \& Future Works}
Unlike previous methods described in the literature, we proposed a simpler \textit{adaptive} model that results in getting faster and reliable results. Also, it can deal with non-stationary nature of anomaly detection problem and adapt itself after observing an anomaly. Moreover, the added object recognition module (Faster R-CNN) improves the true positive with the tradeoff of having trivial false positive. The proposed method has a constraint when there is a strong noise in the input frames as the motion distribution map is built based on the motion velocity of the moving objects. \\
Many different training, tests, and experiments have been left for the future due to lack of time. Future works concern training Faster R-CNN over related anomaly detection datasets. Also, validating the proposed AAD model over longer video sequences.

\bibliographystyle{unsrt}
\bibliography{lib}

%\clearpage
%\section{Team Members}
%\begin{itemize}
%\item Mohammad Farhadi Bajestani
%\begin{itemize}
%\item Implementation of object detection using Faster-RCNN (PYTorch)
%\item Literature Review
%\end{itemize}
%\item Seyed Soroush Heidari Rahmat Abadi
%\begin{itemize}
%\item Implementation Farneback optical flow using OpenCV
%\item Collecting Anomaly Detection Dataset
%\end{itemize}
%\item Seyed Mostafa Derakhshandeh Fard
%\begin{itemize}
%\item Implementing distribution map based on optical flow output
%\item Documentation
%\end{itemize}
%\item Roozbeh Khodadadeh
%\begin{itemize}
%\item Implementing main module for mapping extracted features from object detection and flow %detection to each frame
%\item Documentation
%\end{itemize}

%\end{itemize}

\end{document}